# Land Cover Classification from Remote Sensing Images Based on Multi-Scale Fully Convolutional Network

Rui Li, Shunyi Zheng, Chenxi Duan and Ce Zhang

*Abstract*—**Convolutional neural network (CNN) is an effective method to extract information from remote sensing images for land cover classification. Nevertheless, the frequently-used single-scale convolution kernel limits the scope of information extraction. In this paper, a Multi-Scale Fully Convolutional Network (MSFCN) with a multi-scale convolutional kernel is proposed to exploit discriminative representations from two-dimensional (2D) satellite images. Meanwhile, when it comes to spatio-temporal images, the mainstream 2D fully convolutional neural network (FCN) collapses the temporal dimension when exploiting the spatial features, which ruins the time series information contained in multi-temporal satellite images. Hence, we expand our MSFCN to three-dimension using three-dimensional (3D) CNN, capable of harnessing each land cover category's time series interaction from the reshaped spatio-temporal remote sensing images. What is more, a channel attention block (CAB) and a global pooling module (GPM) are included to enhance the channel consistency and global contextual consistency. Experiments conducted on two spatial datasets and two spatio-temporal datasets both demonstrate the effectiveness of our MSFCN.**

## I. INTRODUCTION

Land cover classification is a foundational technology for land resource management, cultivated area evaluation, and economic assessment, which is significant for homeland security and national economic stability [1]. Conventionally, large-scale field surveys are the primary method to obtain land use and land cover. Although the outcomes of surveys are of high quality, the investigative procedures are time-consuming and labor-intensive. Meanwhile, the information about the geographical distribution of land cover is often missing [2, 3].

As a powerful Earth observation technology, remote sensing can capture Earth's surface images via sensors on aircraft or satellites without physical contact [4]. The optical remote sensing is a significant branch of remote sensing and has been applied in many fields, including super-resolution land cover mapping [5], drinking water protection [6], and object detection [7]. Scholars have increasingly focused on automatic land cover classification using satellite images by profiting from the substantial remote sensing images [8-12].

Generally, remote sensing classification models consist of two procedures, feature engineering and classifier training. The former aims to transform spatial, spectral, or temporal information into discriminative feature vectors. The latter is designed to train a general-purpose classifier to classify the feature vectors into the correct category.

When it comes to land cover classification, vegetation indices are one genre of frequently-used features extracted from multi-spectral/multi-temporal images to manifest physical properties of land cover. The normalized difference vegetation index (NDVI) [13] and soil-adjusted vegetation index (SAVI) [14] highlight vegetation. The normalized difference bareness index (NDBaI) [15] and the normalized difference bare land index (NBLI) [16] emphasize bare land. The normalized difference water index (NDWI) [17] and modified NDWI (MNDWI) [18] indicate water.

Meanwhile, the remote sensing community has tried to design various classifiers from diverse perspectives [19], from orthodox methods such as logistic regression [20], distance measure [21], and clustering [22], to advanced techniques including support vector machine (SVM) [23], random forest (RF) [24], Markov random field (MRF) [25], artificial neural network (ANN) [26], and multi-layer perceptron (MLP) [27]. Since extraction of the geographical distribution of land cover requires pixel-based image classification, precisely refined pixel features are the core of these classifiers. However, the high dependency on manual descriptors restricts the flexibility and adaptability of these methods.

The emergence of Deep Learning (DL), which is powerful to capture nonlinear and hierarchical features automatically, tackles the above deficiency to a great extent. DL has influenced many domains, such as computer vision (CV) [28], natural language processing (NLP) [29], as well as automatic speech recognition (ASR) [30]. As a typical classification task, a great many DL methods have been introduced to land cover

This work was supported in part by the National Natural Science Foundations of China (No. 41671452). *(Corresponding author: Chenxi Duan.)*

R. Li and S. Zheng are with School of Remote Sensing and Information Engineering, Wuhan University, Wuhan 430079, China (e-mail: lironui@whu.edu.cn; syzheng@whu.edu.cn).

C. Duan is with the State Key Laboratory of Information Engineering in Surveying, Mapping, and Remote Sensing, Wuhan University, Wuhan 430079, China; chenxiduan@whu.edu.cn (e-mail: chenxiduan@whu.edu.cn).

C. Zhang is with the Lancaster Environment Centre, Lancaster University, Lancaster LA1 4YQ, United Kingdom (e-mail: c.zhang9@lancaster.ac.uk).

classification. Compared to vegetation indices that only consider finite bands, DL methods can harness various information, including periods, spectrums, and the interactions between different kinds of land cover.

Zhong et al. [31] exploited temporal features using a one-dimensional (1D) CNN to recognize the intricate seasonal dynamics of economic crops and lessened the dependency on hand-crafted feature engineering. Pelletier et al. [8] proposed a temporal CNN for satellite image time series. They proved the significance of harnessing the information both in spectral dimension and temporal dimension when implementing the convolutions. Based on fine-tuned CNN, Tong et al. [32] combined hierarchical segmentation and patch-wise classification for land cover classification. Specifically, many cutting-edge technologies used in semantic segmentation, whose task is assigning each pixel with a specific category [33], have also been generalized to land cover classification. [19]. Inspired by the progress in the encoder-decoder Fully Convolutional Network (FCN) framework, Stoian et al. [34] proposed a Fine-Grained U-Net architecture for sparse annotation images captured by Sentinel-2. Cao et al. [35] incorporated the U-Net and ResNet to classify the tree species using high-resolution images.

Even though the encoder-decoder FCN framework [36-38] has been an essential structure for land cover classification [39-41], the single-scale convolution kernel limits the scope of information extraction. To remedy this drawback, we propose a Multi-Scale Fully Convolutional Network based on encoder-decoder FCN structure to exploit both local and global features from satellite images. We design two branches with convolutional layers in different kernel sizes in each layer of the encoder to capture multi-scale features. Besides, a channel attention block and a global pooling module [42] enhance channel consistency and global contextual consistency.

Meanwhile, spatio-temporal satellite images, bolstered by their increasing attainability, are at the forefront of a comprehensive effort towards automatic Earth monitoring by international agencies [43]. However, when utilizing the 2D CNN to extract features from spatio-temporal satellite images, the temporal dimensions of the extracted features generated by the convolution layer must be averaged and devastated to a scalar, which collapses the time series information contained in multi-temporal images. Many studies have been conducted motivated by NLP's progress to cope with this defect. Rußwurm et al. [44, 45] adapted sequence encoders to represent Sentinel 2 images' temporal sequence and alleviated the demand of humdrum and cumbersome cloud-filtering. Interdonatoa et al. [46] designed a two-branch architecture with an RNN branch to extract temporal features and a CNN branch to extract spatial features. By incorporating both CNN and RNN, Rustowicz et al. [47] designed a 2D U-Net + CLSTM model for spatio-temporal satellite images. Meanwhile, for embedding time-sequences, Transformer architecture was introduced into land cover classification using spatio-temporal satellite images by Garnot et al. [43]. All these attempts have made encouraging progress and broadened the boundaries of this field.

Meanwhile, the advent of 3D CNN solves the dilemma mentioned above from another facet. Unlike traditional 2D CNN, which operates on 2D images, 3D CNN implements convolutional operation on three dimensions, which naturally fits feature extraction from data represented in 3D format. Thus, 3D CNN has been utilized for video understanding [48], point clouds representation [49], 3D object detection based on light detection and ranging (LiDAR) data [50], hyperspectral images classification [51], and multi-temporal images segmentation [52]. As remote sensing images normally comprise much temporal, dynamic, or spectral information, like the whole crop growth cycle in the temporal dimension, 3D CNN is a superexcellent method to extract these features.

Using multi-temporal images, Ji et al. [52] designed a 3D-CNN-based segmentation model for crop classification. As the temporal dimension is reserved, the model's performance surpassed the 2D-CNN-based methods and other traditional classifiers. However, as 3D CNN is a computationally intensive operation, the pixel-by-pixel segmented procedure requires numerous computational resources [52]. Thus, based on the idea of semantic segmentation, Ji et al. [42] proposed a novel 3D encoder-decoder FCN framework with global pooling and attention mechanism (3D FGC), which was able to capture feature maps from the whole input and improves both the accuracy and the efficiency.

Based on the insight and progress mentioned above, we extend our Multi-Scale Fully Convolutional Network to three-dimension based on 3D CNN for land cover classification using spatio-temporal satellite images. To verify the effectiveness, we compare the performance of 2D MSFCN with SegNet [37], FC-DenseNet [53], U-Net [36], Attention U-Net [54] and FGC [42]. and the performance of 3D MSFCN with 1D U-Net, 2D U-Net [36], 3D U-Net [36], Conv-LSTM [44] and 3D FGC [42]. In addition, we expand 2D Attention U-Net [54] to 3D and contrast its capability to MSFCN. The major contributions of this paper could be listed as follows:

1) To expand the scope of information extraction in the spatial domain, we designed a multi-scale convolutional block (MSCB), which can capture the input's local and global features, respectively.

2) Based on MSCB, we proposed a Multi-Scale Fully Convolutional Network (MSFCN) with channel attention block and global pooling module, and extend MSFCN to 3D spatio-temporal satellite images.

3) A series of quantitative experiments on two spatial datasets and two spatio-temporal datasets show the effectiveness of the proposed MSFCN.

This paper's remainder is arranged as follows: In Section 2, taking 3D MSFCN as an example, we illustrate the detailed structure of the proposed framework. The experimental results are provided and analyzed in Sections 3. Finally, in Section 4, we conclude the entire paper.

## II. METHODOLOGY

### A. Feature Extraction using 3D CNN

3D CNN is capable of capturing spatial and temporal features simultaneously, and Batch Normalization (BN) layer [55] is often appended to improve numerical stability. Thus, we consider 3D CNN with a BN layer as an example to elaborate on 3D CNN's mechanism. Supposing that the size of input 3D feature maps is expressed as $(t \times h \times w, c)$, and the shape of the convolution kernel is $(k_t \times k_h \times k_w)$, where $t$, $h$, $w$, and $c$ denote the dimension of time series, height, width, and channels. The convolution operations are implemented between the convolution kernel and sliding windows in the shape of $(k_t \times k_h \times k_w)$. The obtained values constitute the output 3D feature maps. Another important parameter, stride, determines the distance of width and height traversed per slide of the sliding windows. A diagrammatic sketch with one kernel can be seen in Fig. 1. Concretely, the operation of 3D CNN can be formalized as:

$$
\begin{aligned}
x_{i,j}^{t,h,w} &= \sum_m \sum_{p=0}^{T_i-1} \sum_{q=0}^{H_i-1} \sum_{r=0}^{W_i-1} W_{i,j,m}^{p,q,r} x_{i-1,m}^{(t+p),(h+q),(w+r)} \\
&\quad + b_{i,j}
\end{aligned}
\tag{1}
$$

where $x_{i,j}^{t,h,w}$ denotes the $j$th feature cube at position $(t, h, w)$ in the $i$th layer. $m$ means the feature maps generated by the $(i-1)$th layer. $W_{i,j,m}^{p,q,r}$ represents the column weight of the $m$th feature cube at position $(p, q, r)$. $b_{i,j}$ is the $j$th feature cube in the $i$th layer's bias items of the filter. $T_i$ means the convolution kernel along the temporal dimension of input spatio-temporal satellite images, while $H_i$ and $W_i$ respectively express the height and width of the kernel in the spatial dimension.

Then, the generated 3D feature maps $x_i$ is fed into the BN layer and normalized as:

$$
\hat{x}_i = \frac{x_i - E(x_i)}{\sqrt{Var(x_i) + \epsilon}}
\tag{2}
$$

$$
y_i = \sigma(\gamma_i \hat{x}_i + \beta_i)
\tag{3}
$$

where $y_i$ is the output of the BN layer. $Var(\cdot)$ and $E(\cdot)$ represent the variance function and expectation of the input. $\epsilon$ is a small constant to maintain numerical stability. $\gamma$ and $\beta$ are two trainable parameters, and the normalized result $\hat{x}_i$ can be scaled by $\gamma$ and shifted by $\beta$. $\sigma(\cdot)$ denotes the activation function, which is set as ReLU in our model.

As the quality of extracted features limits the performance of the model and the convolution kernel size determines the receptive field, how to design the size of the convolution kernel is the crux of the network.

### B. Multi-Scale Convolutional Block

Generally, the larger convolution kernel size means the larger receptive field and the more global vision, which augments the scope of areas observed in the image. Conversely, the decrease in the convolution kernel size would shrink the receptive field and obtain the local vision. However, both the global visual patterns and the local visual patterns contain visual features. Thus, a fully convolutional neural network's evident imperfection is the same size convolutional kernels, leading to a constant receptive field. As shown in Fig. 2(a), the conventional convolutional block used in FCN usually contains two stacked 3D CNN with the activation function. To expand the receptive field, in MSFCN, we design a multi-scale convolutional block (MSCB) to exploit the global and local features simultaneously.

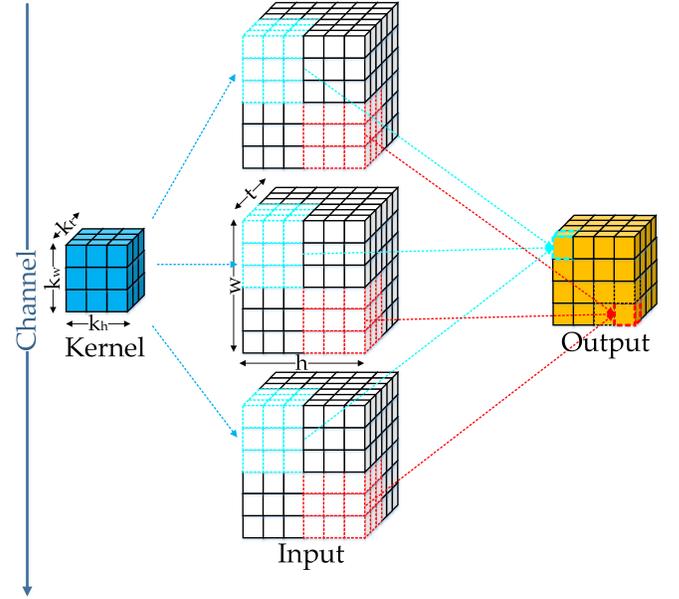

Fig. 1. 3D convolution indicates convolution operator is implemented in three directions (i.e. two spatial directions and a temporal direction) sequentially. Both the input feature maps and the output feature maps are 3D tensors.

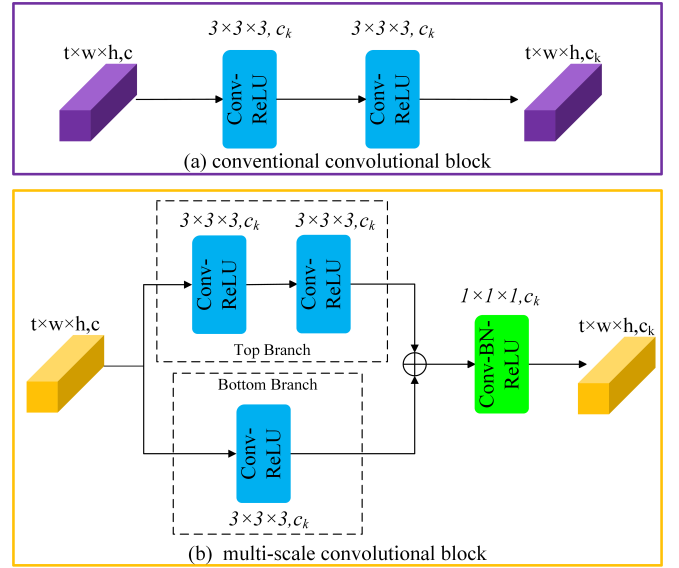

Fig. 2. Comparison of (a) conventional convolution block and (b) multi-scale convolution block.

The structure of the multi-scale fully convolutional layer can be seen in Fig. 2(b). Similarly, supposing the input 3D feature maps is in the shape of $(t \times h \times w, c)$, where the $t$, $h$, $w$, and $c$ represent the time series, height, width, and channels of the input. The top branch of the block contains two stacked $(3 \times 3 \times 3)$ convolution layers, and the receptive field of two stacked $(3 \times 3 \times 3)$ convolution layers are equivalent to a $(5 \times 5 \times 5)$ convolution layer, which can be seen from Fig. 3. Thus, the top branch is capable of capturing more global visual patterns. Meanwhile, the block's bottom branch harnesses a single $(3 \times 3 \times 3)$ convolution layer that exploits local visual patterns.

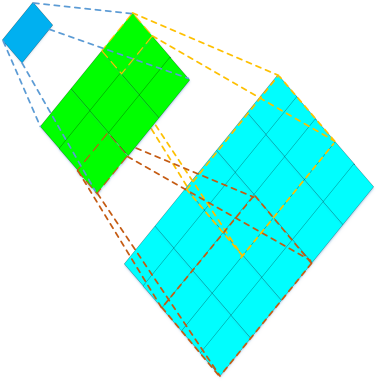

Fig. 3. The receptive field of two stacked (3×3) convolution layers is equivalent to a (5×5) convolution layer, and the same is true of 3D CNN.

Subsequently, the add operation is implemented between the outputs of the top branch and the bottom branch, and obtains the feature maps with the size of $(t \times h \times w, c_k)$. Finally, the extracted feature maps are fed into a $(1 \times 1 \times 1)$ convolution layer with the BN layer to increase further the nonlinear characteristics and characterization capabilities of the block.

### C. Channel Attention Block and Global Pooling Module

In the FCN framework, the convolution operator's output is a score map, which indicates the probability of each class at each pixel. And to attain the final score map, all channels of feature maps are simply summed as:

$$y_n = F(x; \omega) = \sum_{i=1, j=1, k=1}^{D} \omega_{i,j,k} x_{i,j,k} \tag{4}$$

$\omega$ denotes the convolution kernel. $x$ represents the feature maps generated by the network. $D$ is the set of pixel positions. And $n \in \{1, 2, ..., N\}$, where $N$ indicates the number of channels. Then the prediction probability is calculated as:

$$\delta_i(y_n) = \frac{\exp(y_n)}{\sum_{j=1}^{K} \exp(y_j)} \tag{5}$$

where $y$ denotes the output of the network, and $\delta$ indicates the prediction probability. The category with the highest probability is the final predicted label, deduced by Equation 4 and Equation 5. Nevertheless, Equation 4 impliedly demonstrate that all channels share equal weights. However, the features generated by different stages own different levels of discrimination, which causes different consistency in prediction.

Supposing the prediction label is $y_0$ and that the corresponding true label is $y_1$ we can modify the highest probability value from $y_0$ to $y_1$ by introducing a parameter $\alpha$:

$$\bar{y} = \alpha y = \begin{bmatrix} \alpha_1 \\ \vdots \\ \alpha_N \end{bmatrix} \cdot \begin{bmatrix} y_1 \\ \vdots \\ y_N \end{bmatrix} = \begin{bmatrix} \alpha_1 \omega_1 \\ \vdots \\ \alpha_N \omega_N \end{bmatrix} \times \begin{bmatrix} x_1 \\ \vdots \\ x_N \end{bmatrix} \tag{6}$$

in which $\alpha = Sigmoid(x; w)$ and $\bar{y}$ is the new prediction label of the network. As can be seen from Equation 6, the value of $\alpha$ weights the feature maps $x$ and enhances the discriminative features and restrains the indiscriminative features. The channel attention block is designed based on the insight mentioned above [56, 57] and is expanded to the 3D version [42].

The CAB structure can be seen in Fig. 4, whose input is the concatenated feature maps extracted by the encoder and decoder. First, a 3D global average pooling layer in CAB exploits the input's global context, and sequentially two $(1 \times 1 \times 1)$ convolution layers with ReLU and sigmoid activation function adaptively realign the channel-wise dependencies. The weight vector generated by CAB manifests the relative significance between the channel-wise features and enhances the discriminability about features. Subsequently, the multiplication operation and addition operation are operated between the output vector and the input feature maps. Finally, the last $(1 \times 1 \times 1)$ convolution layer is designed to generate globally consistent spatio-temporal feature maps. Through re-modeling the channel-wise features, the 3D channel attention block (CAB) fuses the spatio-temporal features between the encoder and the decoder.

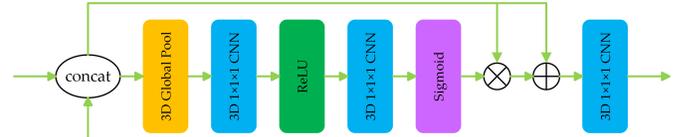

Fig. 4. The structure of the channel attention block (CAB).

Meanwhile, context is utile information that can enhance segmentation and detection performance using deep learning [58]. As for land cover classification, local semantic information contained in per pixel is often equivocal. By taking contextual information into account, the semantic information will be enhanced. Global average pooling is an effective method to capture the global contextual prior [58]. Based on the idea that a global average pooling layer can improve the spatio-temporal consistency on the highest level of the encoder (i.e., the top semantic layer) [57], the global pooling module (GPM) is elaborately designed [42], which can be seen in Fig. 5. Meanwhile, with global spatio-temporal consistency, the GPM transforms the feature maps at the highest level of the encoder to the decoder's corresponding feature maps. Like the CAB, GMP's effect is reweighting feature maps, which can also be seen as an attention mechanism.

The structure of the GMP can be seen in Fig. 5. First, the input feature maps are fed into a $(1 \times 1 \times 1)$ convolution layer. Then, a 3D global average pooling and a $(1 \times 1 \times 1)$ convolution layer with a sigmoid activation function are attached. Finally, the multiplication operation and addition operation are implemented between the generated vector and the first convolution layer's output. The final output is processed by the last $(1 \times 1 \times 1)$ convolution layer to acquire the decoder's highest layer.

## D. Network Architecture

Based on the 3D CNN, the multi-scale convolutional block, the channel attention block, and the global pooling module, we construct the MSFCN for land cover classification from satellite images, as shown in Fig. 6. The encoder of the MSFCN comprises four multi-scale convolutional blocks with the output channels as 32, 64, 128, and 256, respectively, and the number of layers and channels will be discussed in Section III.F. After each multi-scale convolutional block, the max-pooling layer with $(1 \times 2 \times 2)$ kernel is applied, which reserves the temporal information and condenses the spatial information. At the highest layer of the encoder, the GPM is utilized to enhance the global spatio-temporal consistency. Then, using CAB, the feature maps from the encoder and decoder are fused, and the output of each layer in the decoder is sequentially restored up to the input size via the transposed convolution layer with $(1 \times 2 \times 2)$ kernel. After each transposed convolution layer, a $(3 \times 3 \times 3)$ convolution layer is attached to avoid the checkerboard pattern caused by the transposed convolution. In the end, the final 3D feature maps are fed into a $(t \times 3 \times 3)$ convolution layer and a $(1 \times 1 \times 1)$ convolution layer to coalesce time dimension and generate 2D segmentation maps.

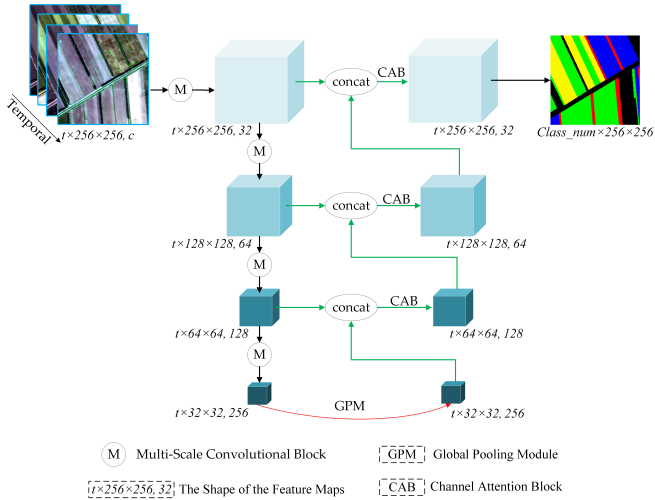

Fig. 6. The structure of the proposed MSFCN network.

The cross-entropy loss function is used as a quantitative evaluation and backpropagation index to measure the disparity between the obtained 2D segmentation maps and ground truth, which is defined as:

$$loss_{i,j} = -\sum_k q_{i,j,k} \log p_{i,j,k} \tag{7}$$

$$loss = \frac{1}{N} \sum_i \sum_j loss_{i,j} \tag{8}$$

where $p_{i,j}$ is the predicted category probability distribution of pixel $(i,j)$, $q_{i,j}$ is the actual category probability distribution of pixel $(i,j)$, $k$ represents the number of classes, and $N$ denotes the number of pixels.

## III. Experimental Results

This section first introduces the datasets and experimental settings to verify the effectiveness of MSFCN and then compares the performance between different frameworks.

### A. Datasets

The effectiveness of 2D MSFCN is verified using Wuhan Dense Labeling Dataset (WHDLD) [59, 60] and Gaofen Image Dataset (GID) [30], which can be seen in Fig.7 and Fig. 8. The effectiveness of 3D MSFCN is verified using two Gaofen 2 (GF2) spatio-temporal satellite images [36], which can be seen in Fig. 9.

WHDLD contains 4940 RGB images in 256 × 256 captured by Gaofen 1 Satellite and ZY-3 Satellite over Wuhan urban area. By image fusion and resampling, the resolution of the images is to reach 2m/pixel. The images contained in WHDLD are labeled with six classes, i.e., bare soil, building, pavement, vegetation, road, and water.

GID contains 150 RGB images in 7200 × 6800 captured by Gaofen 2 Satellite over 60 cities in China. Each image covering a geographic region of $506 \ km^2$. The GID images are labeled with six classes, i.e., build-up, forest, farmland, meadow, water, and others. However, as we don't have enough computing resources to cope with such extremely enormous pixels, we just select 15 images contained in GID. The principle of selection is to cover the whole six classes. And the serial number of the chosen images will be released with our open-source code [1].

The two spatio-temporal satellite datasets that own four bands (red, green, blue, and near-infrared) in 4m ground resolution were gathered in 2015 and 2017, respectively. For the 2015 dataset, there are four images collected in June, July, August, and September in the year of 2015, and 2652 × 1417 pixels of each image. The 2017 dataset comprises seven images with 2102 × 1163 pixels captured in June, July, August, September, October, November, and December in 2017. Two GF2 datasets are preprocessed with the quick atmospheric correction [61] and geometrical rectification.

### B. Experimental Setting

To evaluate the effectiveness of 2D MSFCN, SegNet [37], FC-DenseNet57 (Tiramisu) [53], U-Net [36], Attention U-Net (U-NetAtt) [54] and FGC [42] are taken into comparison. And the performance of 3D MSFCN are compared with 1D U-Net, 2D U-Net [36], 3D U-Net [36], Conv-LSTM [44] and 3D FGC [42]. In addition, we expand 2D Attention U-Net [54] to 3D and contrast its capability with MSFCN.

All the models are implemented with PyTorch, and the optimizer is set as Adam with a 0.0001 learning rate. The batch size is set as 16 for WHDLD and GID, and 4 for GF2 spatio-temporal satellite images. All the experiments are implemented on a single NVIDIA GeForce RTX 2080ti GPU with 11 GB RAM.

For WHDLD, we randomly select 60% images as the training set, 20% images as the validation set, and the rest 20% images as the test set. For GID, we separately partition each image into non-overlap patch sets with the size of 256 × 256 and just discard the pixels on the edges, which cannot be divisible by

---

[1] https://github.com/lironui/Multi-Scale-Fully-Convolutional-Network.

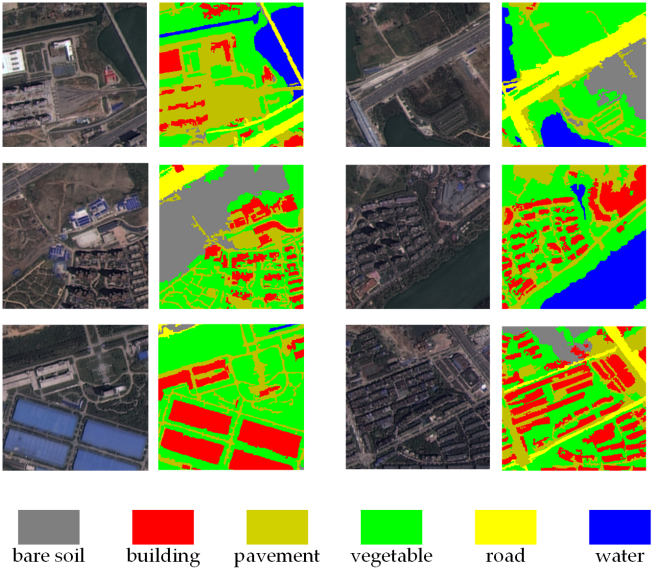

Fig. 7. Examples of WHDLD images and their corresponding ground truth.

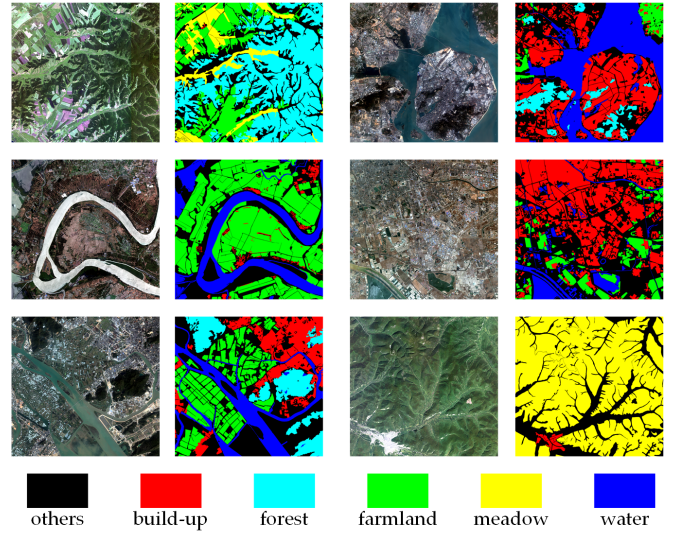

others | build-up | forest | farmland | meadow | water

Fig. 8. Examples of GID images and their corresponding ground truth.

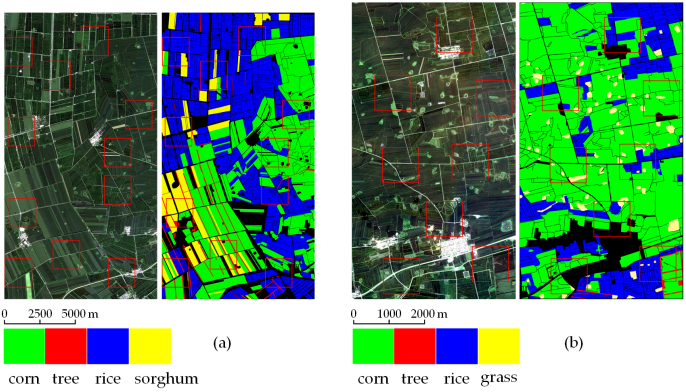

Fig. 9. GF2 datasets gathered in (a) 2015, and (b) 2017. Each dataset owns four crop species labelled in different color, and black pixels represent the label information are absent. Patches indicated in red rectangles were utilized to train the network and the remainder to prediction.

256. Thus, 10920 patches are obtained. We randomly selected 60% patches as the training set, 20% patches as the validation set, and the rest 20% patches as the test set. And the training sets of WHDLD and GID are augmented by horizontal axis flipping, vertical axis flipping, color enhancement, Gaussian blur, and random noise. When training the network, if the accuracy in the validation set is no longer increasing for 10 epochs, we would terminate the training process early to restrain overfitting. The number of training, validation, and test pixels per class for WHDLD and GID is provided in Table I.

For two spatio-temporal satellite images, the samples in each category are severely imbalanced. Thus, we selected a portion of the images that contain samples of all the classes to train the network, indicated in red rectangles in Fig. 9. Since pixels in these two datasets are not abundant, we enlarge the images in the 2015 dataset to the size of 2816 × 1536 and the images in the 2017 dataset to the size of 2304 × 1280 by zero-padding and then segment each image into non-overlap patch sets in the size of 256 × 256 to evaluate prediction accuracy. Of course, the selected portion for training is also set as zero to avoid data leakage. The number of training and test pixels per class is provided in Table II. Each model has trained 100 epochs on the training set and then verified on the test set.

For each dataset, the overall accuracy (OA), average accuracy (AA), Kappa coefficient (K), mean Intersection over Union (mIoU), Frequency Weighted Intersection over Union (FWIoU), and F1-score (F1) are adopted as evaluation indexes. Given the predicted segmentation maps and ground truth, the IoU indicates their intersection size divided by their union size. The mIoU averages the IoU of every category, and the FWIoU weights IoU of each category by the frequency. We select mIoU as the primary indicator, as it reflects both the overall accuracy (OA) and the consistency degree (Kappa) and is becoming a frequently-used indicator for land cover segmentation [41, 62, 63].

### C. Results on WHDLD and GID

The experimental results of different methods on WHDLD and GID are demonstrated in Table III. The performance of the proposed MSFCN transcends other algorithms in all

## TABLE I
### THE SAMPLES FOR EACH CATEGORY FOR TRAINING, VALIDATION AND TEST.

| | | Train | Val | Test |
|---|---|---|---|---|
| WHDLD | bare | 7746403 | 2475482 | 2854410 |
| | building | 21848819 | 7135568 | 6917771 |
| | pavement | 22842445 | 7671979 | 6782834 |
| | road | 8225161 | 2850179 | 2869957 |
| | vegetable | 87444443 | 28505640 | 28859223 |
| | water | 46141433 | 16110720 | 16465373 |
| GID | others | 125858447 | 40426710 | 40061365 |
| | build-up | 49528719 | 16603346 | 17203079 |
| | farmland | 125542298 | 41351598 | 40884984 |
| | forest | 37555494 | 12302122 | 13716761 |
| | meadow | 25657841 | 9335581 | 8437873 |
| | water | 65249073 | 23111267 | 22826562 |

## TABLE II
### THE SAMPLES FOR EACH CATEGORY FOR TRAINING AND TEST.

| 2015 | Train | Test | 2017 | Train | Test |
|---|---|---|---|---|---|
| rice | 253286 | 1069586 | rice | 93931 | 356085 |
| corn | 198585 | 1064487 | corn | 320895 | 1206244 |
| sorghum | 102649 | 193686 | grass | 15140 | 63117 |
| tree | 17410 | 57677 | tree | 3941 | 7787 |

TABLE III
THE EXPERIMENTAL RESULTS ON WHLDL (LEFT) AND GID (RIGHT).

| Method | OA | AA | K | mIoU | FWIoU | F1 | Method | OA | AA | K | mIoU | FWIoU | F1 |
|---|---|---|---|---|---|---|---|---|---|---|---|---|---|
| SegNet | 80.229 | 63.787 | 71.403 | 52.940 | 68.876 | 66.529 | SegNet | 80.035 | 82.396 | 74.612 | 70.962 | 67.420 | 82.290 |
| Tiramisu | 82.188 | 70.712 | 74.903 | 58.167 | 72.243 | 71.276 | Tiramisu | 79.467 | 84.008 | 74.377 | 69.032 | 65.627 | 80.716 |
| U-Net | 81.830 | 67.724 | 74.422 | 55.706 | 72.450 | 68.567 | U-Net | 78.992 | 81.115 | 73.295 | 69.417 | 65.936 | 81.326 |
| U-NetAtt | 82.602 | 69.738 | 75.484 | 56.918 | 73.474 | 69.622 | U-NetAtt | 80.919 | 83.838 | 75.878 | 70.930 | 68.539 | 82.511 |
| FGC | 82.975 | 68.855 | 75.927 | 57.368 | 73.540 | 70.274 | FGC | 81.180 | 84.716 | 76.270 | 72.067 | 68.859 | 83.240 |
| MSFCN | **84.168** | **72.081** | **77.558** | **60.366** | **74.892** | **73.031** | MSFCN | **83.718** | **85.544** | **79.353** | **75.127** | **72.688** | **85.378** |

TABLE IV
PER CLASS F1-SCORE PERFORMANCE ON WHLDL (LEFT) AND GID (RIGHT).

| Method | bare | building | pavement | road | vegetable | water | Method | others | buildup | farmland | forest | meadow | water |
|---|---|---|---|---|---|---|---|---|---|---|---|---|---|
| SegNet | 47.682 | 63.253 | 51.466 | 54.649 | 86.473 | 95.649 | SegNet | 63.451 | 79.085 | 83.510 | 89.241 | 84.962 | 93.493 |
| Tiramisu | 50.313 | 68.918 | 53.576 | **70.047** | 88.206 | 96.598 | Tiramisu | 57.062 | 79.007 | 85.436 | 87.068 | 83.274 | 92.648 |
| U-Net | 43.097 | 70.752 | 52.609 | 58.668 | 89.185 | 97.089 | U-Net | 63.351 | 80.585 | 81.564 | 87.768 | 82.996 | 91.692 |
| U-NetAtt | 47.974 | 72.736 | 48.942 | 60.576 | 89.994 | **97.511** | U-NetAtt | 67.123 | 81.523 | 84.569 | 86.955 | 82.513 | 92.381 |
| FGC | 50.282 | 72.642 | 53.842 | 57.931 | 89.651 | 97.294 | FGC | 66.810 | 81.957 | 84.101 | 89.570 | 84.840 | 92.165 |
| MSFCN | **52.178** | **74.499** | **55.177** | 68.797 | **90.024** | **97.511** | MSFCN | **71.536** | **83.442** | **86.907** | **90.332** | **85.752** | **94.303** |

TABLE VI
THE EXPERIMENTAL RESULTS USING DIFFERENT METHODS ON 2015 DATASET (LEFT) AND 2017 DATASET (RIGHT).

| Method | OA | AA | K | mIoU | FWIoU | F1 | Method | OA | AA | K | mIoU | FWIoU | F1 |
|---|---|---|---|---|---|---|---|---|---|---|---|---|---|
| 1D U-Net | 92.302 | 75.017 | 87.339 | 66.745 | 86.581 | 76.114 | 1D U-Net | 95.709 | 74.331 | 89.365 | 66.091 | 92.065 | 75.924 |
| 2D U-Net | 91.883 | 85.710 | 86.788 | 74.131 | 86.174 | 84.117 | 2D U-Net | 96.369 | 78.015 | 90.933 | 71.873 | 93.491 | 81.449 |
| 3D U-Net | 96.620 | 85.819 | 94.391 | 82.151 | 93.517 | 88.112 | 3D U-Net | 96.662 | 81.836 | 91.851 | 74.375 | 94.252 | 83.497 |
| 3D U-NetAtt | 96.272 | 90.662 | 93.876 | 83.441 | 93.143 | 88.947 | 3D U-NetAtt | 97.102 | 82.320 | 93.020 | 75.505 | 94.904 | 84.38 |
| Conv-LSTM | 96.682 | 90.314 | 94.523 | 84.618 | 93.77 | 91.123 | Conv-LSTM | 96.414 | 81.379 | 91.117 | 75.026 | 93.456 | 84.156 |
| 3D FGC | 96.272 | 90.662 | 93.876 | 83.441 | 93.143 | 90.380 | 3D FGC | 97.083 | 82.052 | 92.841 | 75.387 | 94.767 | 84.311 |
| 3D MSFCN | **97.784** | **93.275** | **96.339** | **87.753** | **95.848** | **92.971** | 3D MSFCN | **97.132** | **85.088** | **93.039** | **77.156** | **94.880** | **86.018** |

TABLE VII
PER CLASS F1-SCORE PERFORMANCE ON 2015 DATASET (LEFT) AND 2017 DATASET (RIGHT)..

| Method | rice | corn | sorghum | tree | Method | rice | corn | grass | tree |
|---|---|---|---|---|---|---|---|---|---|
| 1D U-Net | 97.743 | 92.968 | 75.965 | 37.781 | 1D U-Net | 96.582 | 97.671 | 58.544 | 50.899 |
| 2D U-Net | 97.301 | 92.321 | 72.225 | 74.623 | 2D U-Net | 97.226 | 97.864 | 65.230 | 65.476 |
| 3D U-Net | 98.476 | 95.780 | 82.997 | 75.194 | 3D U-Net | 97.868 | 98.115 | 67.790 | 70.215 |
| 3D U-NetAtt | 98.369 | 97.055 | 92.721 | 67.642 | 3D U-NetAtt | 97.752 | 98.413 | 74.264 | 67.091 |
| Conv-LSTM | 98.733 | 97.154 | 89.791 | 78.813 | Conv-LSTM | 96.643 | 97.940 | 65.798 | **76.244** |
| 3D FGC | 98.670 | 96.839 | 87.997 | 78.013 | 3D FGC | 97.861 | 98.335 | 72.562 | 68.485 |
| 3D MSFCN | **99.184** | **98.203** | **94.317** | **80.180** | 3D MSFCN | **98.236** | **98.586** | **77.660** | 69.589 |

quantitative evaluation indexes, which can be seen from the Table. For WHDLD, the proposed MSFCN brings near 3% improvements both on mIoU and F1-score compared with FGC. And for the GID dataset, the gains are more than 3% in mIoU and more than 2% in F1-score, respectively.

Table IV summarizes the per class F1-score performance of the different methods for WHDLD and GID. The proposed MSFCN obtains the best performance in most classes on WHDLD and whole classes on GID. Meanwhile, we investigate the confusion between each pair of categories, and we report the confusion matrix by heat maps for each competing method in Fig. 10. The more visible diagonal structure (the dark blue blocks concentrated on the diagonal) indicates, the more powerful capacity of distinguishing between classes. And the diagonal structure of MSFCN is more distinct than others, which proves our framework's superiority.

TABLE V
THE COMPARISON OF PARAMETERS AND COMPUTATIONAL COMPLEXITY.

| Method | input shape | Parameters (M) | Complexity (G) |
|---|---|---|---|
| SegNet | | 1.93 | 9.27 |
| Tiramisu | | 29.45 | 40.29 |
| U-Net | 3×256×256 | 1.38 | 11.92 |
| U-NetAtt | | 2.17 | 12.75 |
| FGC | | 2.19 | 8.4 |
| MSFCN | | 2.67 | 9.66 |

The number of parameters and the calculations' consumptions is also significant to assess a framework's merit. The comparison of parameters and computational complexity between different algorithms are reported in Table V, where'M' is the abbreviation of million, the unit of parameter number, and 'G' is the abbreviation of Gillion (thousand million), the unit of floating point operations. And the comparison demonstrates that the design of MSFCN doesn't bring in redundant parameters or lead to high computational complexity.

Some visual results generated by our method and comparisons are provided in Fig. 11.

### D. Results on 2015 and 2017 datasets

To training the network, the inputs of the 1D U-Net are reshaped into ($ct \times 65536$) tensors, and the inputs of the 2D U-Net are reshaped into ($ct \times 256 \times 256$), while the input of the Conv-LSTM, 3D U-Net, 3D FGC, 3D U-NetAtt and 3D MSFCN are ($c \times t \times 256 \times 256$) tensors, where $c$ and $t$ denote the number of spectral channels and time series, respectively.

The experimental results with different methods for two datasets are demonstrated in Table VI. Since 1D CNN's operation destroys both the spatial and temporal dimensions, 1D U-Net's performance is the worst. As 2D CNN's process ruins the temporal dimension when extracting spatio-temporal features, the models based on 3D CNN dramatically outperform the models based on 2D CNN, which prominently demonstrates the superiority of 3D CNN. The performance of Conv-LSTM transcends 2D-based models, as the information contained in the temporal dimension is taken into consideration. Benefitting from the utilization of attention mechanisms, the 3D U-NetAtt performs better than 3D U-Net.

TABLE IX
THE EFFECTIVENESS OF THE MULTI-SCALE CONVOLUTIONAL BLOCK AND ATTENTION MECHANISMS ON WHDLD (LEFT) AND GID (RIGHT).

| Method | OA | AA | K | mIoU | FWIoU | F1 | Method | OA | AA | K | mIoU | FWIoU | F1 |
|--------|------|------|------|------|------|------|--------|------|------|------|------|------|------|
| U-Net | 81.830 | 67.724 | 74.422 | 55.706 | 72.450 | 68.567 | U-Net | 78.992 | 81.115 | 73.295 | 69.417 | 65.936 | 81.326 |
| MSFB | 82.708 | 68.301 | 75.459 | 57.098 | 73.119 | 69.941 | MSFB | 81.579 | 83.429 | 76.620 | 71.992 | 69.715 | 83.276 |
| MSFB+CAB | 83.084 | 70.411 | 76.038 | 58.571 | 73.547 | 71.299 | MSFB+CAB | 82.675 | 84.693 | 78.111 | 73.672 | 70.987 | 84.321 |
| MSFB+GPM | 83.433 | 70.214 | 76.608 | 58.608 | 74.347 | 71.003 | MSFB+GPM | 82.891 | 84.136 | 78.302 | 73.671 | 71.575 | 84.453 |
| MSFCN | 84.168 | 72.081 | 77.558 | 60.366 | 74.892 | 73.031 | MSFCN | 83.718 | 85.544 | 79.353 | 75.127 | 72.688 | 85.378 |

TABLE X
THE EFFECT CAUSED BY THE NUMBER OF LAYERS (LEFT) AND THE NUMBER OF CHANNELS (RIGHT) ON GID.

| Method | OA | AA | K | mIoU | FWIoU | F1 | Method | OA | AA | K | mIoU | FWIoU | F1 |
|--------|------|------|------|------|------|------|--------|------|------|------|------|------|------|
| MSFCN3 | 79.513 | 80.862 | 74.178 | 69.858 | 67.243 | 81.723 | MSFCNN | 80.218 | 83.652 | 70.530 | 70.530 | 67.529 | 82.104 |
| MSFCN4 | 83.718 | 85.544 | 79.353 | 75.127 | 72.688 | 85.378 | MSFCN | 83.718 | 85.544 | 79.353 | 75.127 | 72.688 | 85.378 |
| MSFCN5 | 84.449 | 86.554 | 80.300 | 76.042 | 73.843 | 86.010 | MSFCNW | 84.352 | 86.966 | 80.230 | 75.669 | 73.365 | 85.733 |

Similarly, FGC's performance exceeds U-Net due to the consistency enhanced by CAB and GPM. Our proposed MSFCN obtains the state-of-the-art mIoU on two datasets, as the well-designed multi-scale convolutional blocks capture both the global and local features. Table VII reports the per class F1-score performance of the different methods for the 2015 dataset and 2017 dataset. The proposed MSFCN obtains the best performance in whole classes on the 2015 dataset and most classes on the 2017 dataset. The confusion matrix reported by heat maps for each competing method is provided in Fig. 12. And Fig. 13 demonstrates the segmentation maps on two datasets. The first three rows are from the 2015 dataset, and the remainder is from the 2017 dataset. Taking the fourth column as an example, the proposed MSFCN differentiates corn (green) and grass (yellow) better than other models.

Table VIII provides the number of parameters, and the consumption of calculation, which illustrates the complexity of the proposed MSFCN is not unacceptable.

TABLE VIII
THE COMPARISON OF PARAMETERS AND COMPUTATIONAL COMPLEXITY.

| Method | input shape | Parameters (M) | Complexity (G) |
|--------|-------------|----------------|----------------|
| 1D U-Net | 16×65536 | 3.74 | 24.16 |
| 2D U-Net | 16×256×256 | 10.86 | 14.18 |
| 3D U-Net | | 4.87 | 74.69 |
| 3D U-NetAtt | | 5.67 | 121.74 |
| Conv-LSTM | 4×4×256×256 | 0.30 | 77.31 |
| 3D FGC | | 5.32 | 78.51 |
| 3D MSFCN | | 6.58 | 91.46 |

### E. Effectiveness of the Multi-Scale Convolutional Block and Attention Mechanisms

We verify the effectiveness of the multi-scale convolutional block and attention mechanisms in this section. Concretely, we analyzed the proposed MSFCN without multi-scale convolutional block (MSFB), channel attention block (CAB), and global pooling module (GPM) both on WHDLD and GID. The results are shown in Table IX.

The 3D U-Net obtains mIoU of 0.55706 and 0.69417 on WHDLD and GID. By utilizing multi-scale convolutional blocks, the mIoUs reach to 0. 57098, and 0.71992. And the introduction of channel attention block and global pooling module brings 0.01473/0.01510 for WHDLD and 0.01680/0.01679 for GID improvements on mIoU, respectively. The mIoUs are further improved to 0.60366 and 0.75127 when all blocks are introduced.

### F. Investigation about the Number of Layers and Channels

The number of layers and channels are two vital parameters that impact the model's performance and determine the computational complexity. Thus, it is worthwhile to investigate the influence of the number of layers and channels.

Therefore, we implement experiments to inquire about the effect caused by the number of layers. Concretely, we design an MSFCN with 3 layers (MSFCN3) and an MSFCN with 5 layers (MSFCN5) and compare their performance with the MSFCN with proposed 4 layers MSFCN (MSFCN4). As finite layers limit the capacity of representations, the performance of MSFCN3 is significantly weaker than MSFCN4. Specifically, without enormous increases in the parameters and computational complexity, MSFCN4 surpasses MSFCN3 more than 5% on mIoU, seen from Table X (left). However, notwithstanding certain improvements boosted by MSFCN5, the number of parameters of MSFCN5 is four times more than MSFCN4's (seen from Table XI), which is not an efficient option.

TABLE XI
THE COMPARISON OF PARAMETERS AND COMPUTATIONAL COMPLEXITY.

| Method | input shape | Parameters (M) | Complexity (G) |
|--------|-------------|----------------|----------------|
| MSFCN3 | | 2.52 | 6.77 |
| MSFCN4 | 3×256×256 | 2.67 | 9.66 |
| MSFCN5 | | 10.73 | 12.55 |
| MSFCNN | | 0.67 | 2.46 |
| MSFCN | 3×256×256 | 2.67 | 9.66 |
| MSFCNW | | 10.65 | 38.24 |

Besides, we design experiments to research the impact caused by the number of channels. Specifically, we design a narrow MSFCN (MSFCNN) with [16, 32, 64, 128] channels, and a wide MSFCN (MSFCNW) with [64, 128, 256, 512] channels, and compare their performance with the proposed MSFCN with [32, 64, 128, 256] channels. The results from Table X (right) show that the performance of MSFCN surpasses MSFCNN near 5% on mIoU. Meanwhile, with five times on parameters and computational complexity, MSFCNW just brings nearly a 1% improvement.

Based on the above experiments, we can conclude that the proposed MSFCN delicately balances performance and complexity.

## IV. CONCLUSION

In this paper, to implement land cover classification using satellite images, we propose a Multi-Scale Fully Convolutional Network (MSFCN). Firstly, multi-scale convolutional blocks are elaborately designed to expand the scope of information extraction in the spatial domain, capturing both the satellite images' local and global information. Secondly, a channel attention block and a global pooling module enhance channel consistency and global contextual consistency. Thirdly, we extend MSFCN to 3D for spatio-temporal satellite images based on 3D CNN to replace 2D FCN, which adequately utilizes each land cover class's time series interaction on the temporal dimension.

Experiments on two spatial datasets provide the effectiveness of the proposed MSFCN. And experiments on two spatio-temporal datasets demonstrate the 3D CNN is a valid method to exploit information from spatio-temporal images. Meanwhile, we explore the impact of the number of layers and channels, which may provide useful references for designing a land cover classification network based on FCN.

## V. DATA AVAILABILITY STATEMENT

The data used to support the findings of this study are included within the article.

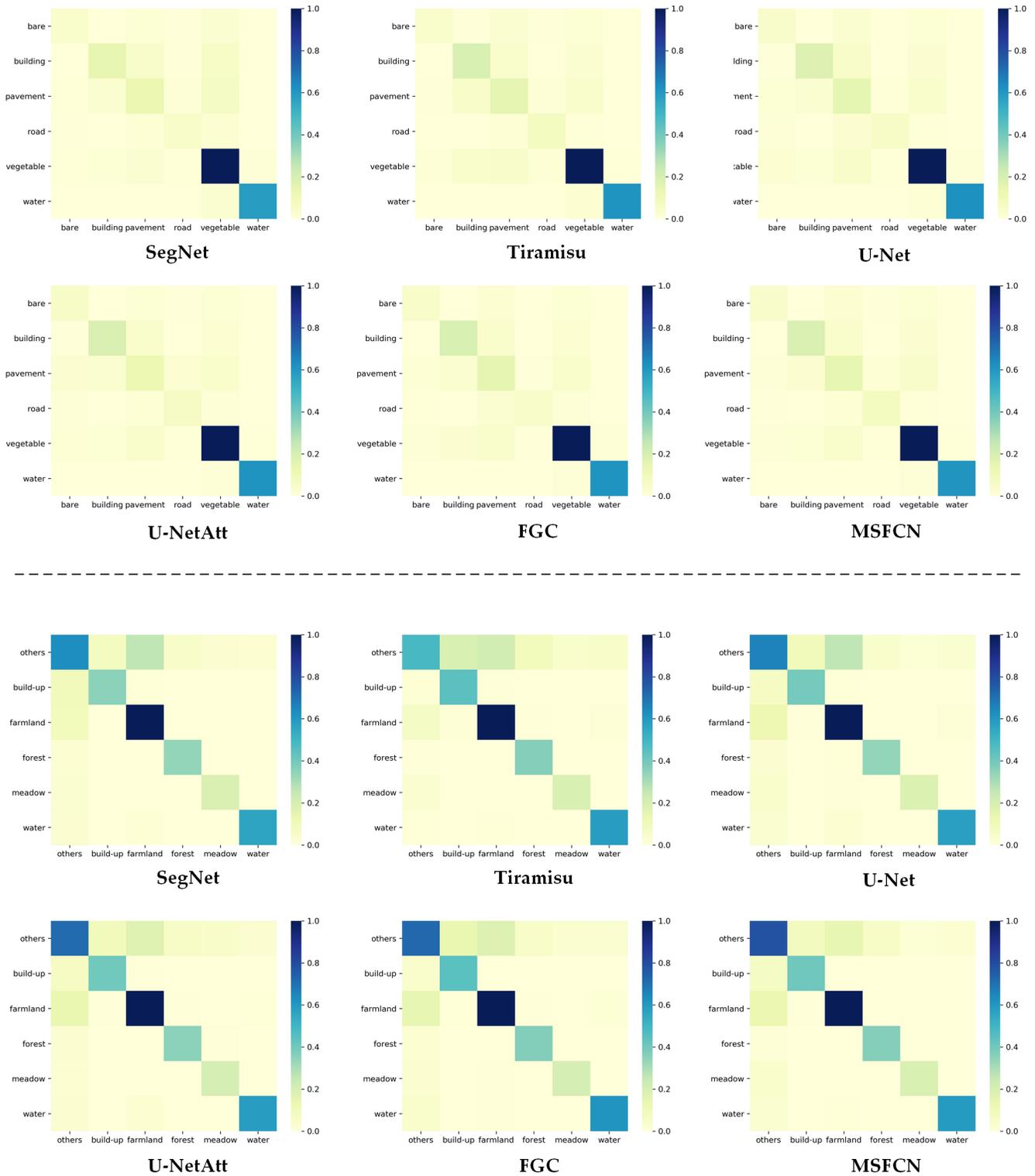

Fig. 10. Heat Maps of different methods on WHDLD (TOP) and GID (BOTTOM).

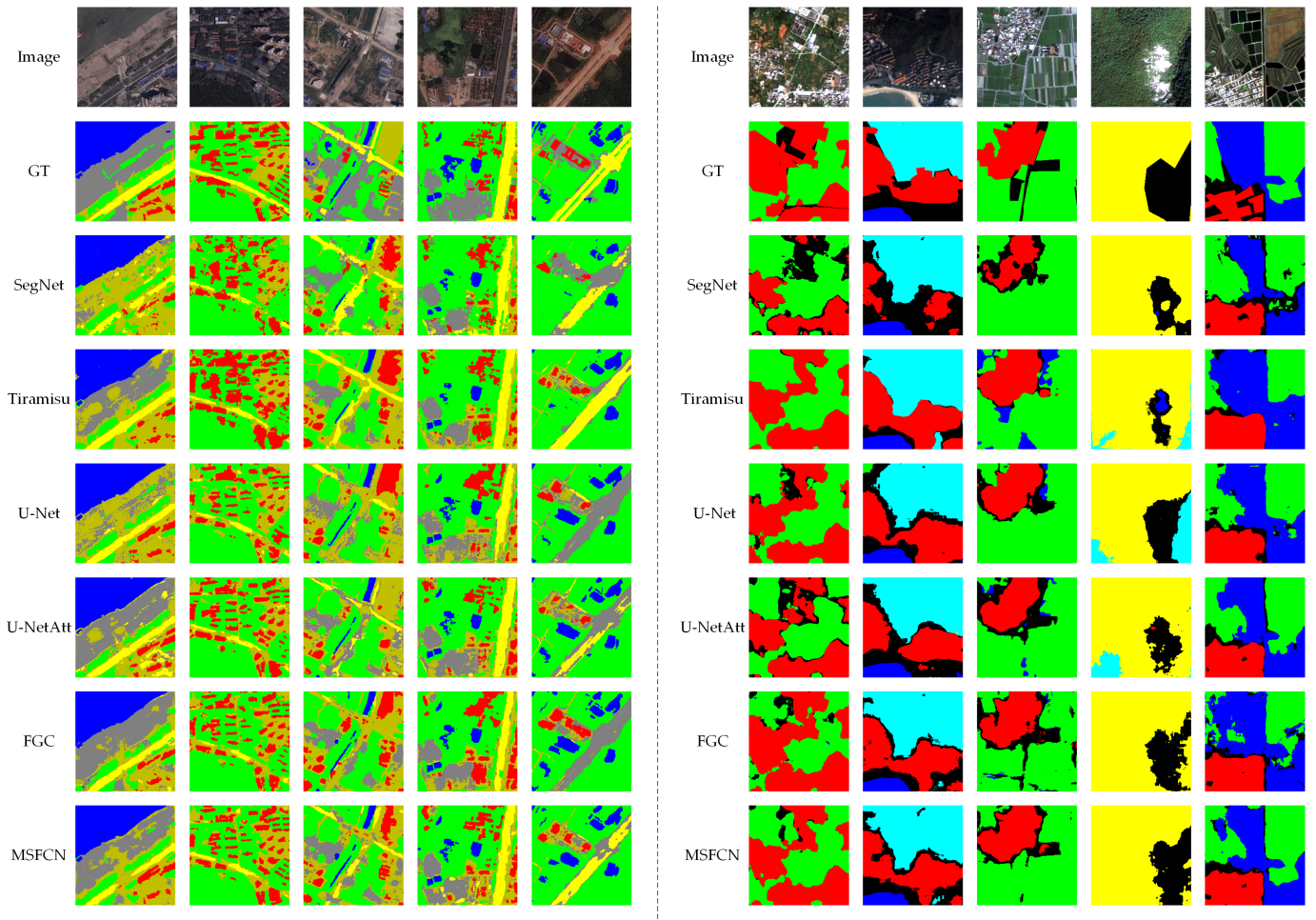

Fig. 11. Land cover classification results of the method proposed and comparisons on WHDLD (LEFT) and GID (RIGHTs).

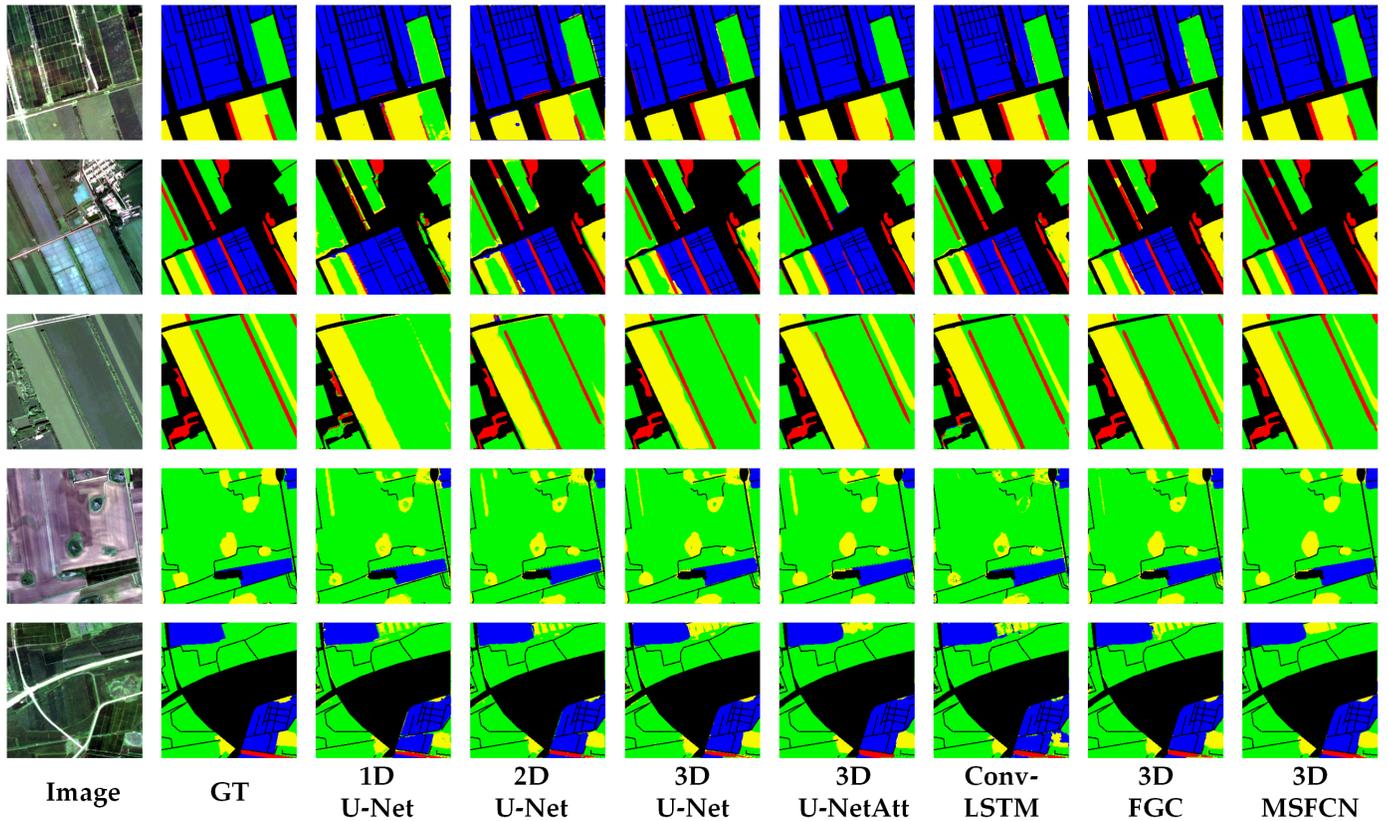

Fig. 13. Land cover classification results of the method proposed and comparisons on 2015 dataset and 2017 dataset.

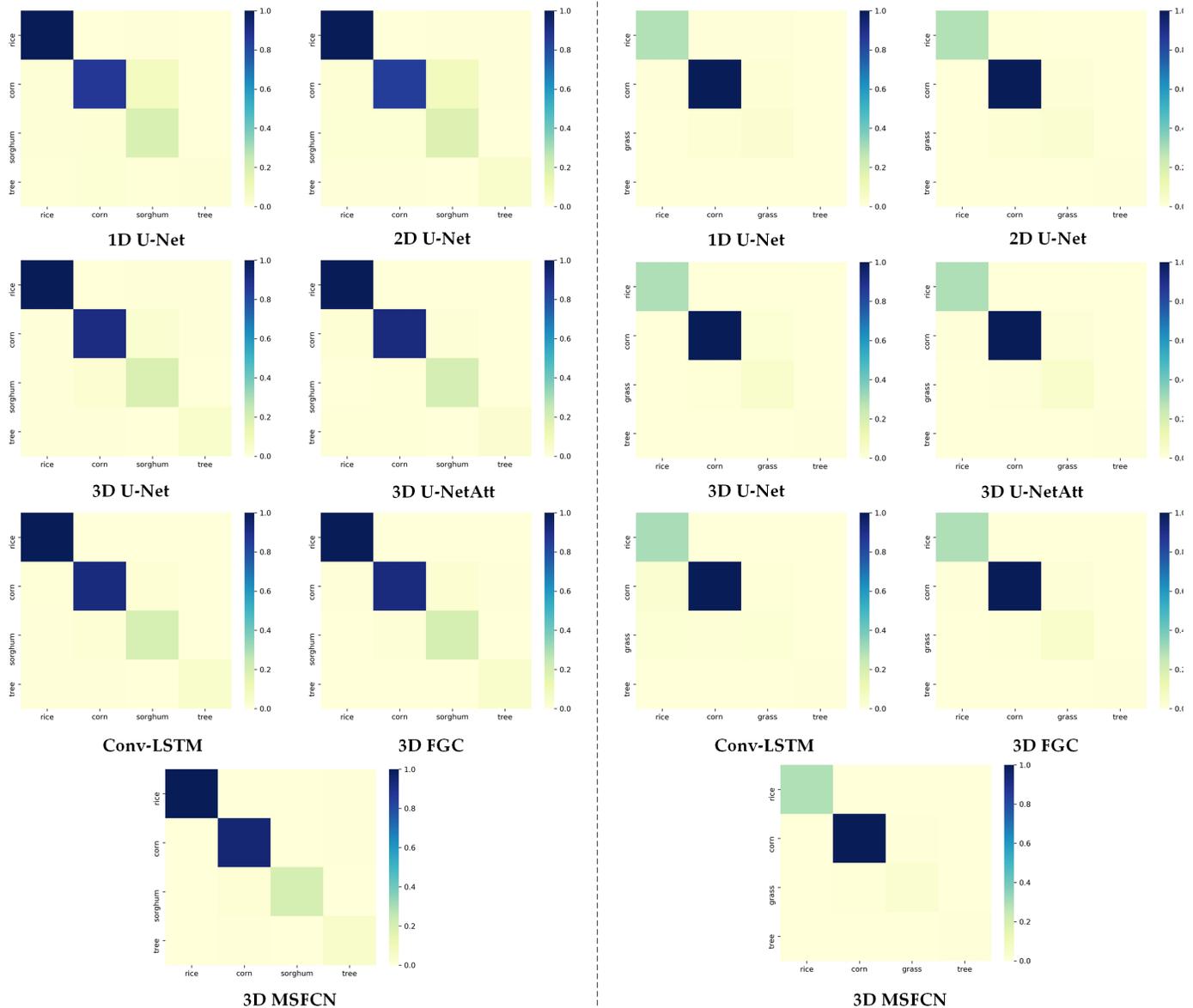

Fig. 12. Heat Maps of different methods on 2015 (LEFT) and 2017 (RIGHT) datasets.